\documentclass{article}
\PassOptionsToPackage{numbers, compress}{natbib}

\usepackage[preprint]{neurips_2025}

\usepackage[utf8]{inputenc} 
\usepackage[T1]{fontenc}    
\usepackage{hyperref}       
\usepackage{url}            
\usepackage{booktabs}       
\usepackage{amsfonts}       
\usepackage{nicefrac}       
\usepackage{microtype}      
\usepackage{xcolor}         
\usepackage{amsmath}
\usepackage{booktabs} 
\usepackage{caption}  
\usepackage{makecell} 
\usepackage{siunitx}  
\usepackage{multirow}
\usepackage{graphicx}
\usepackage{subfigure}
\usepackage{geometry}

\usepackage{hhline} 
\usepackage{natbib}
\usepackage{color}
\usepackage{xcolor,soul}
\usepackage{colortbl}
\definecolor{lightorange}{rgb}{0.98, 0.75, 0.5}
\definecolor{oorange}{RGB}{215,122,71}
\definecolor{yyellow}{RGB}{230,185,79}
\definecolor{ppurple}{RGB}{122,30,97}
\definecolor{ggreen}{RGB}{112,173,71}
\definecolor{lightblue}{rgb}{0.85, 0.95, 0.99}
\definecolor{battleshipgrey}{rgb}{0.3, 0.3, 0.3}
\definecolor{brilliantrose}{rgb}{1.0, 0.33, 0.64}
\definecolor{americanrose}{rgb}{1.0, 0.01, 0.24}
\definecolor{jweigreen}{rgb}{0,0.45,0.24}
\definecolor{jweired}{rgb}{0.45,0,0}
\definecolor{yellowish}{rgb}{0.8, 0.8, 0}
\definecolor{ao(english)}{rgb}{0.0, 0.5, 0.0}
\definecolor{blanchedalmond}{rgb}{1.0, 0.92, 0.8}
\definecolor{atomictangerine}{rgb}{1.0, 0.6, 0.4}
\definecolor{chocolate(web)}{rgb}{0.82, 0.41, 0.12}
\definecolor{bananayellow}{rgb}{1.0, 0.88, 0.21}
\definecolor{goldenbrown}{rgb}{0.6, 0.4, 0.08}
\definecolor{aliceblue}{rgb}{0.94, 0.97, 1.0}
\definecolor{beige}{rgb}{0.96, 0.96, 0.86}
\definecolor{babyblue}{rgb}{0.54, 0.81, 0.94}
\definecolor{camel}{rgb}{0.76, 0.6, 0.42}
\definecolor{cinnamon}{rgb}{0.82, 0.41, 0.12}
\definecolor{darkgreen}{rgb}{0.0, 0.4, 0.2}
\definecolor{darkred}{rgb}{0.55, 0.0, 0.0}
\definecolor{redlinkcolor}{rgb}{0.79607843, 0.25098039, 0.25882353}
\definecolor{bluecitecolor}{rgb}{0,0.36,0.69}

\usepackage[normalem]{ulem} 
\usepackage{bbm}
\useunder{\uline}{\ul}{}
\newcommand{\ie}{i.e., }
\newcommand{\eg}{e.g., }

\newcommand*{\affaddr}[1]{#1}
\newcommand*{\affmark}[1][*]{\textsuperscript{#1}}

\title{\raisebox{-0.3\height}{\includegraphics[scale=0.08]{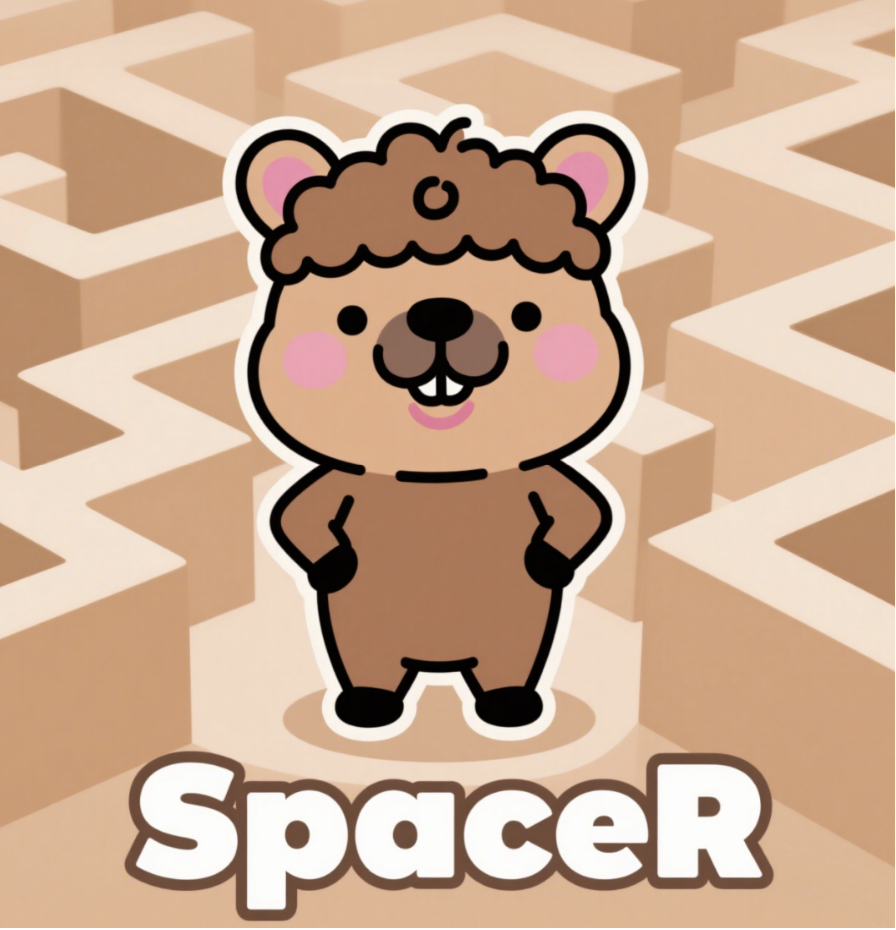}}\hspace{0.2em}SpaceR: Reinforcing MLLMs in Video Spatial Reasoning}

\author{
  Kun Ouyang\affmark[1] \quad
  Yuanxin Liu\affmark[1] \quad
  Haoning Wu\affmark[2] \quad
  Yi Liu\affmark[1] \quad \\
  \textbf{Hao Zhou}\affmark[3] \quad 
  \textbf{Jie Zhou}\affmark[3] \quad
\textbf{Fandong Meng}\affmark[3] \quad
  \textbf{Xu Sun}\affmark[1]\thanks{Corresponding Author(s)} \\
  \affaddr{\affmark[1] National Key Laboratory for Multimedia Information Processing,\\
      School of Computer Science, Peking University}\\
  \affaddr{\affmark[2] Nanyang Technological University} \quad
  \affaddr{\affmark[3] WeChat AI, Tencent Inc., China} \\
  \texttt{kunouyang10@gmail.com} \quad \texttt{xusun@pku.edu.cn} \\
}

\begin{document}

\maketitle

\begin{abstract}
Video spatial reasoning, which involves inferring the underlying spatial structure from observed video frames, poses a significant challenge for existing Multimodal Large Language Models (MLLMs). This limitation stems primarily from 1) the absence of high-quality datasets for this task, and 2) the lack of effective training strategies to develop spatial reasoning capabilities. Motivated by the success of Reinforcement Learning with Verifiable Reward (RLVR) in unlocking LLM reasoning abilities, this work aims to improve MLLMs in video spatial reasoning through the RLVR paradigm. To this end, we introduce the \textbf{SpaceR} framework. First, we present \textbf{SpaceR-151k}, a dataset with 91k questions spanning diverse spatial reasoning scenarios with verifiable answers, and 60k samples for maintaining general multimodal understanding. Second, we propose \textbf{Spatially-Guided RLVR (SG-RLVR)}, a novel reinforcement learning approach that extends Group Relative Policy Optimization (GRPO) with a novel map imagination mechanism, which encourages the model to infer spatial layouts in the thinking process, thereby facilitating more effective spatial reasoning.
Extensive experiments demonstrate that SpaceR achieves state-of-the-art performance on spatial reasoning benchmarks (e.g., VSI-Bench, STI-Bench, and SPAR-Bench), while maintaining competitive results on video understanding benchmarks (e.g., Video-MME, TempCompass, and LongVideoBench). 
Remarkably, SpaceR surpasses the advanced GPT-4o by 11.6\% accuracy on VSI-Bench and is on par with the leading proprietary model Gemini-2.0-Flash, highlighting the effectiveness of our SpaceR-151k dataset and SG-RLVR in reinforcing spatial reasoning ability of MLLMs. Code, model, and dataset are available at~\url{https://github.com/OuyangKun10/SpaceR}. 
\end{abstract}

\begin{figure}[h]
    \centering
    \includegraphics[width=0.8\textwidth]{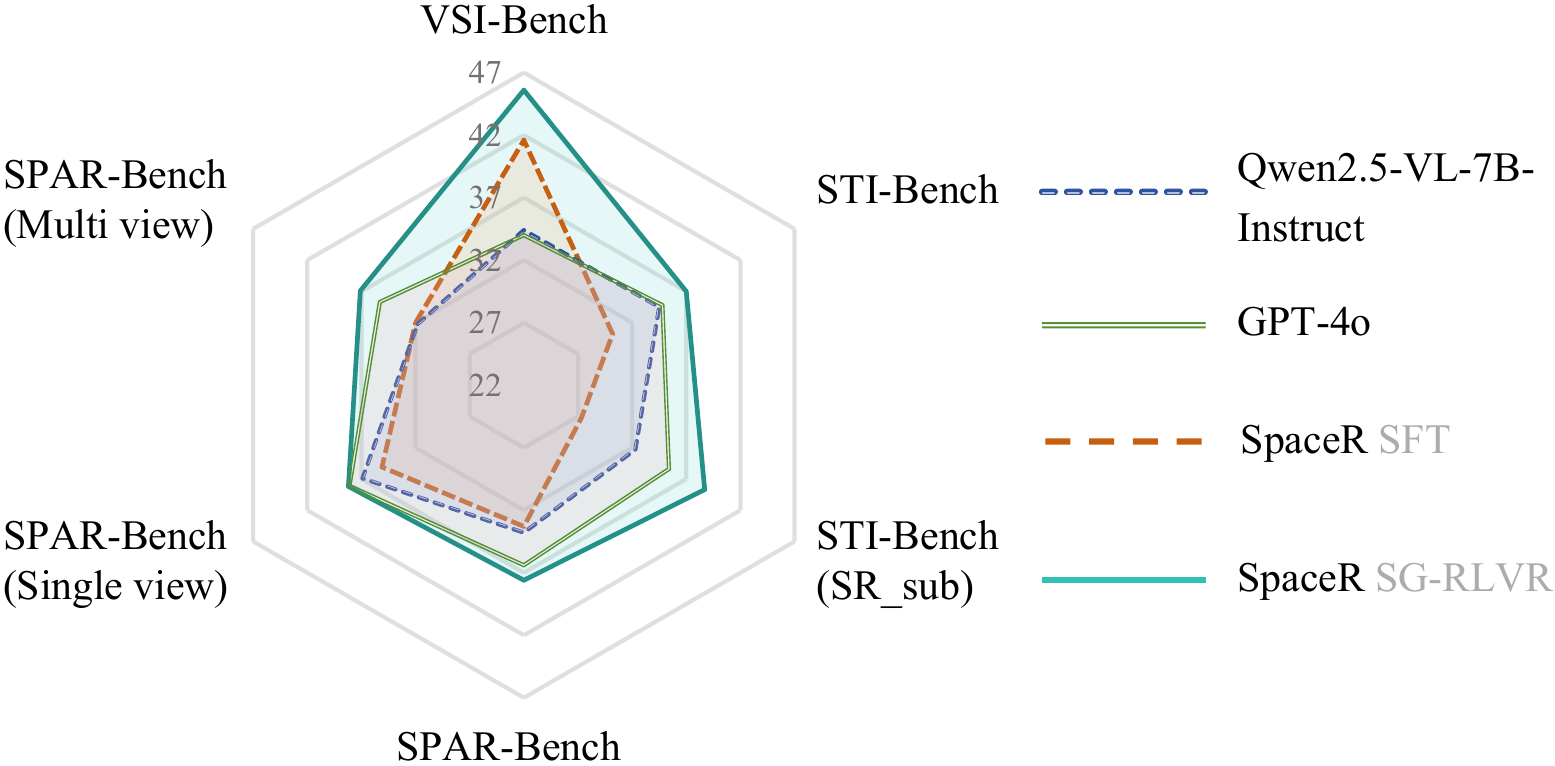}
     \caption{Performance comparison on spatial reasoning benchmarks and their corresponding subsets for Qwen2.5-VL-7B-Instruct~\cite{Qwen2.5-VL}, GPT-4o~\cite{GPT-4o}, our SpaceR \textcolor{gray}{SFT} and SpaceR \textcolor{gray}{SG-RLVR}.} 
    \label{Fig:radar}
\end{figure}
\section{Introduction}\label{main_intro}

Video spatial reasoning~\cite{yang2024thinking} requires reconstructing 3D spatial layouts from sequences of observed frames. This task demands a higher-level cognitive ability than conventional video understanding tasks, such as video captioning~\cite{venugopalan2015translating}, video question answering~\cite{vqaiccv}, and temporal grounding~\cite{gao2017talltemporalactivitylocalization}, which typically require only recall of video content. Although recent advancements in Multimodal Large Language Models (MLLMs) have significantly improved performance on conventional video understanding~\cite{chen2024internvl,Qwen2.5-VL,GPT-4o}, these models still struggle with video spatial reasoning~\cite{li2025stibenchmllmsreadyprecise,yang2024thinking}. This limitation stems mainly from two factors: 1) the absence of a high-quality dataset specifically designed for spatial reasoning, and 2) the reliance of most existing MLLMs on supervised fine-tuning (SFT) during post-training, which is insufficient for fostering deep reasoning capabilities.

In contrast to SFT, recent studies have demonstrated that Reinforcement Learning with Verifiable Rewards (RLVR) is more effective in enhancing the reasoning capabilities of both LLMs and MLLMs within pure-text \cite{deepseekai2025deepseekr1incentivizingreasoningcapability} and multimodal tasks \cite{du2025virgopreliminaryexplorationreproducing,feng2025videor1reinforcingvideoreasoning,liu2025visualrftvisualreinforcementfinetuning}. For example, DeepSeek-R1-Zero~\cite{deepseekai2025deepseekr1incentivizingreasoningcapability} utilizes the Group Relative Policy Optimization (GRPO) algorithm to unlock the reasoning capabilities of LLMs. Concurrent work Video-R1~\cite{feng2025videor1reinforcingvideoreasoning} introduce a large-scale multimodal understanding dataset Video-R1-260k and extend GRPO with a novel temporal reward for RLVR, which also achieves promising performance in video understanding benchmarks like MVBench~\cite{li2023mvbench}. Inspired by these findings, this work aims to advance MLLMs' spatial reasoning abilities in video through RLVR.

Specifically, we propose the \textbf{SpaceR} framework, which encompasses two key innovations: \textbf{First}, we introduce the SpaceR-151k dataset, which consists of 151k samples, including 91k spatial reasoning QA pairs (SR-91k) curated based on a 3D reconstruction dataset ScanNet~\cite{dai2017scannet}, and 60k samples drawn from the general multimodal understanding dataset Video-R1-260k~\cite{feng2025videor1reinforcingvideoreasoning}. In particular, SR-91k spans six spatial reasoning tasks (e.g., relative direction, object/room size, and appearance order), filling a critical gap in available resources. \textbf{Second}, we extend the GRPO~\cite{shao2024deepseekmath} paradigm to enhance spatial reasoning by designing task-specific verifiable rewards for various QA formats (\eg multiple choice, numerical). Furthermore, we design a novel map imagination mechanism, where models are prompted to generate an optional cognitive map, a structured representation of object positions in space, within specialized tags \texttt{<map>}$\cdots$\texttt{</map>}. And a map reward is employed to evaluate the quality of these inferred spatial layouts, which incentivizes models to think in space deeply.

Extensive experiment demonstrate that our SpaceR delivers consistent and significant performance gains across several challenging spatial reasoning benchmarks, including VSI-Bench~\cite{yang2024thinking}, STI-Bench~\cite{li2025stibenchmllmsreadyprecise}, and SPAR-Bench~\cite{zhang2025flatland}, while maintaining promising results in representative video understanding benchmarks like Video-MME~\cite{fu2024video}, TempCompass~\cite{liu-etal-2024-tempcompass}, and LongVideoBench~\cite{wu2024longvideobench}. Notably, our model achieves $45.6\%$ accuracy on VSI-Bench~\cite{yang2024thinking}, 
outperforming the advanced proprietary model GPT-4o~\cite{GPT-4o} by 11.6\% accuracy, which is presented in Figure~\ref{Fig:radar}. These empirical results validate both the utility of the SpaceR-151k dataset and the effectiveness of our SG-RLVR in unlocking spatial reasoning capabilities of MLLMs.

Our contributions are threefold.
\begin{itemize}
    \item 
    We introduce the SpaceR-151k dataset specifically designed for video spatial reasoning. It consists of questions spanning diverse spatial reasoning scenarios with verifable answers, addressing the scarcity of resources in this domain.
    \item 
    We propose SG-RLVR, a spatially-guided reinforcement learning framework that integrates a novel map imagination mechanism. It encourages the model to explicitly generate spatial layouts to facilitate video spatial reasoning.
    \item We conduct extensive evaluations across spatial reasoning and video understanding benchmarks, demonstrating that SpaceR achieves state-of-the-art spatial reasoning capabilities and promising generalizability in video understanding, validating the effectiveness of both the SpaceR-151k dataset and our SG-RLVR framework.
\end{itemize}

\section{Related Works}
\subsection{Video Spatial Reasoning}
Video understanding tasks like video captioning~\cite{venugopalan2015translating}, temporal grounding~\cite{gao2017talltemporalactivitylocalization,jin2022embracing}, and temporal perception~\cite{patraucean2023perception,lin2023diversifying}, primarily focus on recalling or summarizing video content. For instance, video captioning necessitates models to generate relevant textual descriptions of the video based on human prompts. Recent advances in Multimodal Large Language Models (MLLMs) have significantly improved performance on these tasks.
Unlike these conventional understanding tasks, video spatial reasoning requires models not only to perceive visual content but also to infer and reconstruct the spatial structure of entire scenes. It is worth noting that video spatial reasoning is crucial for the development of world models~\cite{liu2025worldmodelmillionlengthvideo} and embodied agents~\cite{10.5555/3618408.3618748}. Recent studies~\cite{li2025stibenchmllmsreadyprecise,yang2024thinking,zhang2025flatland} have highlighted the persistent shortcomings of MLLMs on spatial reasoning, underscoring the need for further research in this area. A key limitation contributing to this gap is the scarcity of high-quality training data specifically tailored for video spatial reasoning, which we aim to address in this work.
\subsection{Reinforcement Learning with Verifiable Reward}
Recent works like o1~\cite{openai2024openaio1card}, DeepSeek-R1~\cite{deepseekai2025deepseekr1incentivizingreasoningcapability}, Kimi k1.5~\cite{kimiteam2025kimik15scalingreinforcement} have demonstrated significant breakthroughs in enhancing the reasoning capabilities of large language models (LLMs) through Reinforcement Learning (RL). In particular, the Group Relative Policy Optimization (GRPO)~\cite{shao2024deepseekmath} algorithm, applied in DeepSeek-R1, has revealed the strong potential of Reinforcement Learning with Verifiable Reward (RLVR) framework in equipping MLLMs with advanced reasoning capacity. Building on this foundation, several subsequent efforts~\cite{liu2025visualrftvisualreinforcementfinetuning,peng2025lmm} have employed RLVR to boost visual reasoning performance. For example, Visual-RFT~\cite{liu2025visualrftvisualreinforcementfinetuning} improved MLLMs in multimodal detection~\cite{DBLP:conf/iccv/PlummerWCCHL15}, grounding~\cite{DBLP:conf/eccv/YuPYBB16}, and classification~\cite{DBLP:journals/ijcv/RussakovskyDSKS15}.
LMM-R1~\cite{peng2025lmm} empowers 3B MLLMs with strong reasoning abilities of mathematics through two-stage rule-based RL. Nevertheless, research specifically targeting video spatial reasoning remains underexplored. Motivated by this gap, our work seeks to design an effective reasoning paradigm to enhance MLLMs' capabilities in video spatial reasoning.
\begin{figure*}[htb]
    \centering
    \includegraphics[width=0.98\textwidth]{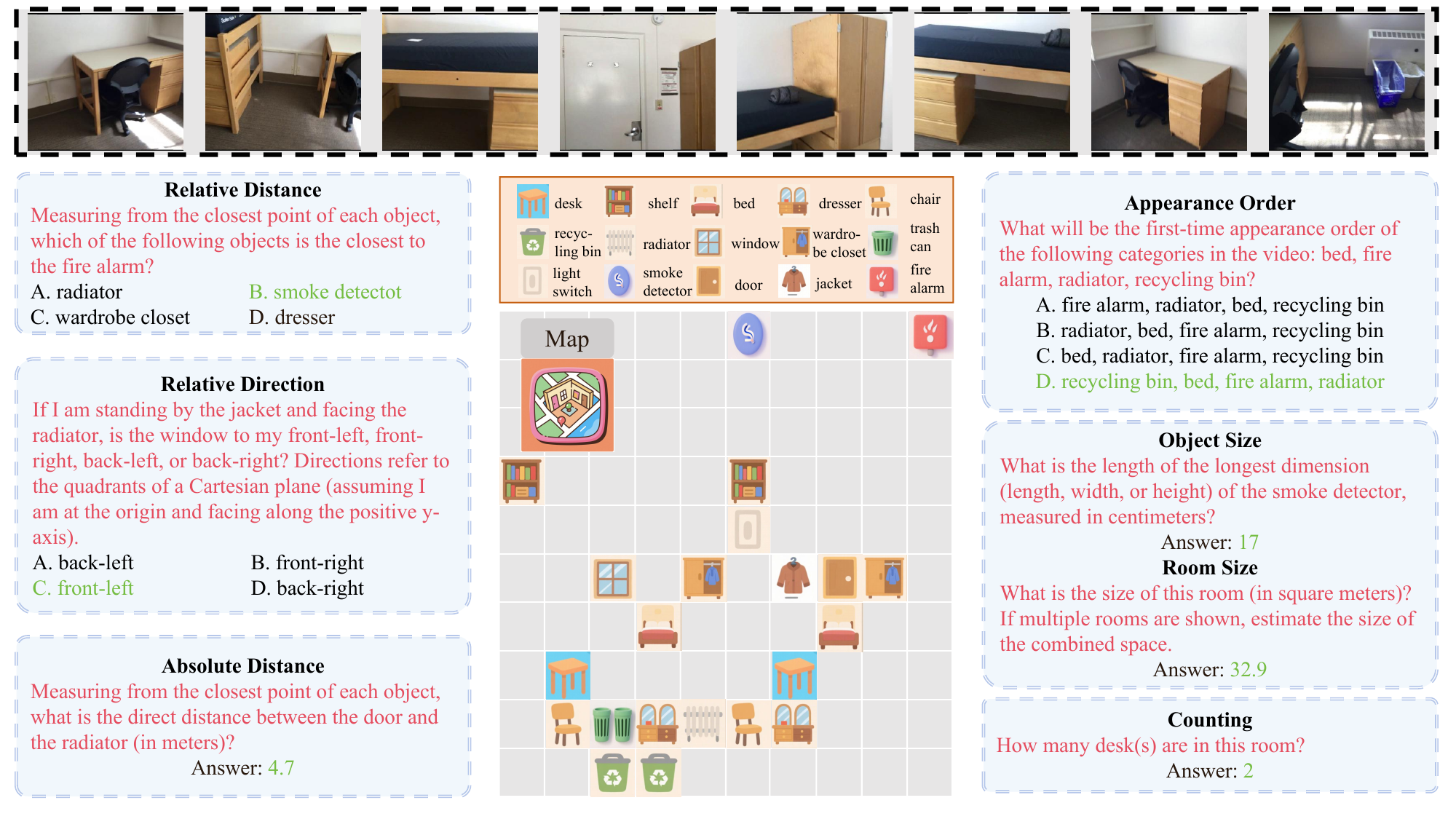}
     \caption{The overview of the Question-Answering examples from SR-91k, including multi-choice QA (\eg relative distance, relative direction, appearance order) and numerical QA (\eg object/room size, absolute distance, and counting), as well as the corresponding map for the video.}
    \label{Fig:QA}
\end{figure*}

\section{Dataset Construction}\label{main_data_construction}
\begin{figure*}[h]
    \centering
    \includegraphics[width=0.95\textwidth]{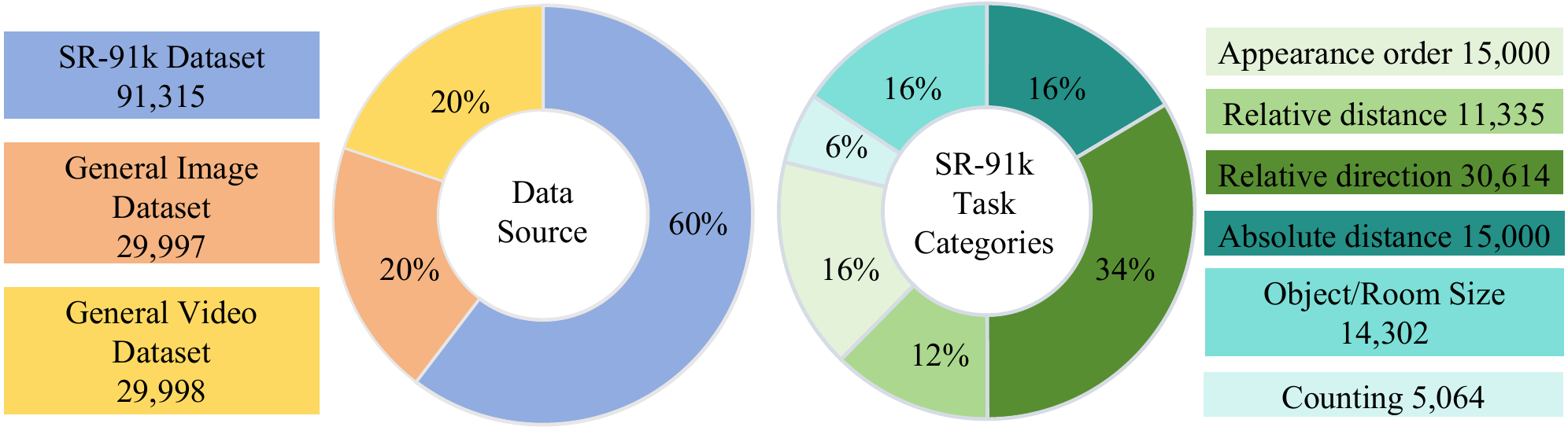}
     \caption{Data statistics of our SpaceR-151k. Left: the distribution of data sources. Right: the task category distribution within SR-91k.}
    \label{Fig:data_distribution}
\end{figure*}

To address the scarcity of high-quality data for video spatial reasoning and maintain general video understanding in the meanwhile, we construct \textbf{SpaceR-151k}, a large-scale dataset consisting of two parts: 1) \textbf{SR-91k}, a tailored spatial reasoning dataset built upon the 3D indoor scene reconstruction dataset ScanNet~\cite{dai2017scannet}, and 2) 60k QA instances resampled from the general multimodal understanding dataset Video-R1-260k~\cite{feng2025videor1reinforcingvideoreasoning}.
The construction process follows three stages: data collection, QA generation, and data filtering. An overview of QA types and examples is provided in Figure~\ref{Fig:QA}, while data statistics are presented in Figure~\ref{Fig:data_distribution}.

\noindent\textbf{Data Collection.}
1) Spatial reasoning. We first parse ScanNet into a unified meta-information format, including object categories, appearance indices, bounding box and other relevant attributes, to facilitate QA generation.
RGB frames of ScanNet are resampled at $24$ FPS to form video clips. Besides, we construct a $10 \times 10$ map for each video to summarize the object distribution of the room, which is exemplified in Figure~\ref{Fig:QA}. Each object's coordinate is determined by the center point of its bounding box and projected onto a 2D map, which is represented in a dictionary format for structured spatial reference.
2) General understanding. 
To preserve general comprehension capabilities, we uniformly sample $60,000$ diverse QA instances from Video-R1-260k. This subset covers multiple QA types including multi-choice, numerical, OCR, free-form, and regression.

\noindent\textbf{QA Generation.}
Leveraging the parsed meta-information of ScanNet, we automatically generate the question-answering (QA) pairs for spatial reasoning tasks, which are categorized into multi-choice QA (\eg relative distance, relative direction, and appearance order) and numerical QA (\eg object/room size, absolute distance, and counting). The QA examples are presented in Figure~\ref{Fig:QA}.
\begin{itemize}
    \item \textbf{Relative Distance}. For each video, we first identify unique objects, randomly select a target object and four candidate objects to be incorporated into the question template. Finally, the minimum Euclidean distance between each target and candidate is computed to determine the answer.

    \item \textbf{Relative Direction}. Utilizing previous identified unique objects in the video, we randomly select three of them to be integrated in the question template. Relative directions are determined on the basis of their center points of bounding box.

    \item \textbf{Appearance Order}.
      We record the first frame index where each object appears, and randomly sample four objects from them to generate the questions. The ordering is determined by their first frame indices.

    \item \textbf{Object/Room Size}.
        Object size is defined as the longest dimension of an object computed from point clouds, and is converted to centimeters. Room size (in square meters) is estimated via the Alpha Shape algorithm\footnote{\url{https://en.wikipedia.org/wiki/Alpha_shape}.}.
    
    \item \textbf{Absolute Distance}. 
    We uniformly sample points within the object bounding boxes and estimate the minimum Euclidean distance between two unique objects in the video.

    \item \textbf{Counting}. 
    We obtain the number of each object appearing in the video from the meta-information of ScanNet.

\end{itemize}

\noindent\textbf{Data Filtering.}
To ensure the quality of the spatial reasoning QA pairs, we apply a series of filtering steps.
First, we limit the number of QA pairs per video to promote scene diversity. For multi-choice QA, we randomly shuffle the positions of correct answers to balance answer distribution and eliminate position bias. In addition, we meticulously adjust the numerical value distribution in the numerical QA to prevent skewed or unrealistic value shifts. After filtering, we retain 91k high-quality QA pairs, forming the SR-91k dataset for spatial reasoning.

\noindent\textbf{Data Statistics.}
The data statistics of SpaceR-151k are exhibited in Figure~\ref{Fig:data_distribution}. This dataset comprises a total of $151,310$ samples, integrating 91k spatial reasoning QA pairs (SR-91k) with 60k instances drawn from the Video-R1-260k. 
SpaceR-151k features a diverse range of QA types: multi-choice, numerical, OCR, free-form, and regression, whose answers are verifiable. Above all, SpaceR-151k provides rich sources for both spatial reasoning and general understanding, which is the foundation of subsequent training. More details for dataset construction and statistics can be found in Appendix~\ref{Appendix_dataconstruction}.
\section{Spatially-Guided Reinforcement Learning with Verifiable Reward}\label{main_training}
To reinforce video spatial reasoning in MLLMs, we propose a reinforcement learning framework named SG-RLVR, which builds on Group Relative Policy Optimization (GRPO)~\cite{deepseekai2025deepseekr1incentivizingreasoningcapability} by introducing verifiable reward functions tailored to diverse QA types and a novel map imagination mechanism to guide spatial reasoning.
\subsection{Verifiable Reward Function}
To supervise model outputs across multiple QA types, including multi-choice, numerical, OCR, free-form, and regression, we design a set of verifiable reward functions that assess either response format or correctness based on task-specific criteria.

\noindent\textbf{Format Reward.} To ensure the model responses adhere to a predefined structure, we define a format reward $R_{\text{format}}$ based on whether the model wraps its reasoning process and answer within \texttt{<think>}$\cdots$\texttt{</think>} and \texttt{<answer>}$\cdots$\texttt{</answer>} tags, respectively: 
\begin{equation} \label{eq:fmt_reward}
R_{format}(\hat{y}) = 
\begin{cases} 
1, & \text{if $\hat{y}$ matches format}, \\
0, & \text{otherwise}.
\end{cases}
\end{equation}

\noindent\textbf{Multi-choice Reward.} For multi-choice QA, the reward $R_{\text{mc}}$ is binary, based on exact match with the ground truth: 
\begin{equation} \label{eq:multi-choice_reward}
R_{mc}(\hat{y}, y) = 
\begin{cases} 
1, & \text{if } \hat{y} = y, \\
0, & \text{otherwise},
\end{cases}
\end{equation}
where $\hat{y}$ is the model's response and $y$ is the ground truth.

\noindent\textbf{Numerical Reward.} To assess numerical values, we compute relative accuracy across varying confidence thresholds $\theta_i \in \{0.5, 0.55, \dots, 0.95\}$. The numerical reward $R_{num}$ is defined as: 
\begin{equation} \label{eq:numerical_reward}
R_{num}(\hat{y}, y) = \frac{1}{N} \sum_{i=1}^{N} \mathbbm{1} \left( \frac{\left|\hat{y} - y\right|}{y} \leq 1 - \theta_i \right),
\end{equation}
$N$ is the number of confidence thresholds.
Besides, for general multimodal understanding data from Video-R1-260k, we incorporate three additional reward functions: OCR, free-form, and regression rewards~\cite{feng2025videor1reinforcingvideoreasoning}. The OCR reward is computed based on the Word Error Rate (WER)\footnote{\url{https://en.wikipedia.org/wiki/Word_error_rate}}, which measures the edit distance between the predicted and reference text. The free-form reward is calculated as the average of ROUGE-1, ROUGE-2, and ROUGE-L scores~\cite{Lin2004ROUGEAP} between the model’s response and the ground truth. The regression reward is determined by the relative distance between the numerical values of response and ground truth.

\subsection{Map-Based Spatial Reasoning }
Since previous RLVR frameworks like GRPO~\cite{deepseekai2025deepseekr1incentivizingreasoningcapability} lack explicit reward signals for spatial information comprehension when applied to video spatial reasoning, we propose a map imagination mechanism that encourages the model to think in space. Specifically, the model is guided to generate a $M \times M$ map to identify object distributions within the scene, supporting downstream reasoning and lead to more reliable answer. To evaluate the quality of the generated map, we design a novel map reward $R_{map}$ that provides precise quantitative feedback to facilitate spatial reasoning.
Particularly, we first calculate the relative accuracy between predicted object and ground truth object by their relative distance $\frac{\sqrt{(x_{p,i} - x_{g,i})^2 + (y_{p,i} - y_{g,i})^2}}{\sqrt{M^2 + M^2}}$, where $M$ is the size of grid map, and average the relative accuracy across all objects to derive map reward $R_{map}$.
Mathematically,

\begin{equation} \label{eq:map_reward}
R_{map}= \sum_{i=1}^{k} \left( \frac{n_i}{\sum_{j=1}^{k} n_j} \times \left( 1 - \frac{\sqrt{(x_{p,i} - x_{g,i})^2 + (y_{p,i} - y_{g,i})^2}}{\sqrt{M^2 + M^2}} \right) \right),
\end{equation}
where $k$ is the number of object categories, $n_i$ is the number of $i$-th object. $\left(x_{p,i}, y_{p,i}\right)$ and $\left(x_{g,i}, y_{g,i}\right)$ are the coordinates of the $i$-th object in the predicted map and ground truth map. 
To regulate the reasoning process, we introduce a length-based reward $R_l$ that encourages outputs to fall within a defined length range: $[l_{min},l_{max}]$. This helps strike a balance between promoting sufficient reasoning and avoiding overthinking. $R_l$ is applied only when the model produces a correct answer within the desired length.
Formally, the map imagination augmented reward $R_m$ is defined as:
\begin{equation}\label{eq:map_aug_reward}
R_m = 
\begin{cases} 
R_{format} + R_{task} + R_{map} + R_{l}, & \text{if } R_{task} = 1 \\
R_{format} + R_{task} + R_{l}, & \text{otherwise}, 
\end{cases}
\end{equation}
where $task \in \{mc, num, ocr, free, reg\}$. This reward shaping ensures that when the model answers correctly and can properly understand the space of the indoor scene, it receives additional reward, pushing the optimization toward adopting a more spatial aware reasoning policy.
The advantage $A_i$, representing the relative quality of the $i$-th response $o_i$, is computed over the updated rewards within each group of responses $[o_1, o_2, \cdots, o_G]$, where $G$ is the number of output responses. The final optimized policy $\pi_\theta$ is prevented from deviating far from the original model parameters $\pi_{\text{ref}}$ by adding a KL-divergence term $\mathcal{D}_{\text{KL}}\left( \cdot \| \cdot \right)$ to the following formulation:

\begin{equation}\label{eq:advantage}
A_i = \frac{R_m - \operatorname{mean}(\{R_m\})}{\operatorname{std}(\{R_m\})}.
\end{equation}
The final policy update follows the clipped surrogate objective of GRPO:

\begin{equation}
\begin{split}
J(\theta) = \mathbb{E}_{q,\{o_i\}} \Biggl[ \frac{1}{G} \sum_{i=1}^{G} \Biggl( & \min \left( \frac{\pi_\theta(o_i|q)}{\pi_{\theta_{\text{old}}}(o_i|q)} A_i, \operatorname{clip}\left( \frac{\pi_\theta(o_i|q)}{\pi_{\theta_{\text{old}}}(o_i|q)}, 1-\epsilon, 1+\epsilon \right) A_i \right) \Biggr) \\
& - \beta \mathcal{D}_{\text{KL}}(\pi_\theta \| \pi_{\text{ref}}) \Biggr],
\end{split}
\end{equation}
where $\beta$ is a regularization coefficient, preventing excessive deviation from the reference policy during optimization, $\epsilon$ is a positive coefficient limits the policy updating degree.

\section{Experiment}\label{main_exp}

\subsection{Experimental Setups}
\label{main_exp_set}
\noindent\textbf{Implementation Details.}
1) In the training stage, we adopt Qwen-2.5-VL-7B-Instruct\footnote{\url{https://huggingface.co/Qwen/Qwen2.5-VL-7B-Instruct}.} as the base model.
The training process is conducted for a maximum of 2 epochs with a per-device batch size of 1. $8$ response candidates are generated for each sample. The maximum completion length is set to $1,024$ tokens. $l_{min}$ and $l_{max}$ are set to $360$ and $512$, respectively. To balance computational efficiency and model performance, we restrict the number of video frames to 16, with each frame processed at a resolution of $128 \times 28 \times 28$.
2) In the evaluation period, we prompt SpaceR to explicitly perform a step-by-step reasoning process on spatial reasoning benchmarks, while directly generating answers for video understanding benchmarks. For the base model, we prompt it to directly produce answers, as it is not trained to perform intermediate reasoning.
The generation temperature is uniformly set to $0.01$. The maximum number of new tokens is set to $1,024$ when reasoning steps are included, and $128$ for direct answer generation. The number of video frames is standardized to 32 during evaluation, and each frame is processed at a resolution of $448 \times 28 \times 28$.

\noindent\textbf{Benchmarks.}
A diverse set of evaluation benchmarks is employed to comprehensively assess the model's capabilities in both spatial reasoning and video understanding. We conduct extensive evaluation on three spatial reasoning benchmarks (\ie VSI-Bench~\cite{yang2024thinking}, STI-Bench~\cite{li2025stibenchmllmsreadyprecise}, and SPAR-Bench~\cite{zhang2025flatland}) and three video understanding benchmarks (\ie Video-MME~\cite{fu2024video}, TempCompass~\cite{liu-etal-2024-tempcompass}, and LongVideoBench~\cite{wu2024longvideobench}).
The detailed description and usage for each benchmark, as well as the evaluated baselines, are summarized in Appendix~\ref{Appendix_exp}.

\begin{table*}[ht]
    \centering
    \resizebox{\textwidth}{!}{
        \begin{tabular}{l|c|c|c|cc|ccc|c|c|c}
        \toprule

            & \multicolumn{1}{c|}{\multirow{3}{*}{\#Params}}& \multicolumn{1}{c|}{\multirow{3}{*}{Frames}} & \multicolumn{6}{c|}{\textbf{Spatial Reasoning}} & \multicolumn{3}{c}{\textbf{Video Understanding}} \\ \cline{4-12}

             &&  & \multicolumn{1}{c|}{\multirow{2}{*}{\textbf{VSI-Bench}}}&\multicolumn{2}{c|}{\textbf{STI-Bench}} & \multicolumn{3}{c|}{\textbf{SPAR-Bench}} & \multicolumn{1}{c|}{\multirow{2}{*}{\textbf{VM}}} & \multicolumn{1}{c|}{\multirow{2}{*}{\textbf{TC}}} & \multicolumn{1}{c}{\multirow{2}{*}{\textbf{LV}}} \\ \cline{5-9}
             & & & & Overall & SR\_sub & Overall & Single-view & Multi-view & & & \\
        \midrule
            \rowcolor{lightgray}\multicolumn{12}{l}{\textit{Closed-source Models}} \\

            GPT-4o~\cite{GPT-4o} &-&-&34.0&34.8&35.4&36.4&38.1&35.3&71.9&73.8&66.7\\
            Gemini 1.5 Pro&-&-&48.8&-&-&-&-&-&75.0&67.1&64.0\\
            Gemini 2.0 Flash &-&-&45.4 &38.7&39.8&-&-&-&-&-&-\\
            Gemini 2.5 Pro &-&-&-&40.9&40.5&-&-&-&-&-&- \\
              \midrule
          \rowcolor{lightgray}\multicolumn{12}{l}{\textit{Open-source Models}} \\
          VideoLLaMA3-7B~\cite{damonlpsg2025videollama3}&7B &- &-&26.9&27.2&-&-&-&66.2&68.1&59.8  \\

          LLaVA-OneVision-7B~\cite{li2024llava}&7B&-&32.4&-&-&31.2&33.1&29.9&58.2&-&56.3\\
        MiniCPM-V-2.6~\cite{yao2024minicpm} &8B&-&- &26.9&29.6&-&-&-&60.9&-&54.9\\ 
          Kimi-VL-A3B-Instruct~\cite{kimiteam2025kimivltechnicalreport} &3B/16B  &16 &37.4 &-&-&-&-&-&62.3&70.3&58.0 \\
        InternVL2.5-78B~\cite{chen2024expanding}&78B&- &-&28.4&29.8&-&-&-&72.1&-&63.6\\
          Qwen2.5-VL-72B-Instruct~\cite{Qwen2.5-VL} &72B &32 &35.6&40.8&36.9&36.4&40.6&33.6&61.3&75.3&57.1\\

          \rowcolor{aliceblue}Qwen2.5-VL-7B-Instruct~\cite{Qwen2.5-VL} & 7B & 32 & 34.4 & 34.5 & 32.3 & 33.8 & 36.9 & 31.8 & 56.3 & 71.1& 53.5 \\
          \midrule
         SpaceR \textcolor{gray}{SFT}&7B&32&41.6&30.2&27.3&33.3&35.1&32.0&57.6&69.3&54.3\\
            \rowcolor{lightblue} SpaceR \textcolor{gray}{SG-RLVR}& 7B & 32     &{45.6}~\textcolor{jweigreen}{ \small($\uparrow$ 11.2)}  & {37.0}~\textcolor{jweigreen}{ \small($\uparrow$ 2.5)}  & {38.7}~\textcolor{jweigreen}{ \small($\uparrow$ 6.4)} & {37.6}~\textcolor{jweigreen}{ \small($\uparrow$ 3.8)} & {38.2}~\textcolor{jweigreen}{ \small($\uparrow$ 1.3)} & {37.1}~\textcolor{jweigreen}{ \small($\uparrow$ 5.3)}&57.9~\textcolor{jweigreen}{ \small($\uparrow$ 1.6)}&71.4~\textcolor{jweigreen}{ \small($\uparrow$ 0.3)}&54.6~\textcolor{jweigreen}{ \small($\uparrow$ 1.1)}\\
        \bottomrule
        \end{tabular}
    }
    \caption{Evaluation results of base model \colorbox{aliceblue}{Qwen2.5-VL-7B-Instruct}, \colorbox{lightblue}{SpaceR}, and other baselines on spatial reasoning benchmarks (VSI-Bench, STI-Bench, and SPAR-Bench), and video understanding benchmarks: \textbf{VM} (Video-MME), \textbf{TC} (TempCompass), and \textbf{LV} (LongVideoBench). SR\_sub is a subset containing six spatial reasoning sub-tasks of STI-Bench.}
    \label{Tab:main_exp}
\end{table*}

\subsection{Main Results}
The evaluation results on the six benchmarks are presented in Table~\ref{Tab:main_exp}. And we have the following observations and analyses.

\noindent\textbf{Overall Analysis.}
Overall, our SpaceR consistently outperforms the base model Qwen2.5-VL-7B-Instruct across all benchmarks.
In spatial reasoning benchmarks, SpaceR even surpasses the proprietary GPT-4o model, highlighting its superior spatial reasoning capabilities. Notably, SpaceR achieves a significant improvement in accuracy gains $11.2$ on VSI-Bench, a representative benchmark for spatial reasoning, underscoring its enhanced reasoning ability to model complex spatial relationships.
Beyond spatial reasoning, SpaceR also generalizes well to video understanding tasks, achieving higher accuracy across all three benchmarks: Video-MME, TempCompass, and LongVideoBench, compared to Qwen2.5-VL-7B-Instruct. This indicates that the spatial reasoning enhancements and general multimodal understanding training samples contribute to broader video comprehension capabilities.

\noindent\textbf{SG-RLVR vs SFT.} 
We further compare the effectiveness of our proposed SG-RLVR and SFT. While SFT yields localized improvements on benchmarks, such as VSI-Bench, Video-MME, and LongVideoBench, it leads to performance degradation on other benchmarks, indicating limited generalizability. In contrast, SG-RLVR consistently improves performance across both spatial reasoning and video understanding benchmarks, highlighting its better generalizability. These results support the claim that ``SG-RLVR generalizes, while SFT memorizes,'' establishing SG-RLVR as a more effective training paradigm for enhancing spatial reasoning in MLLMs.

\begin{figure*}[htb]
       \centering
        \subfigure[Sample distribution.]{
            \label{Fig:data_sampling1}
            \includegraphics[width=.45\linewidth]{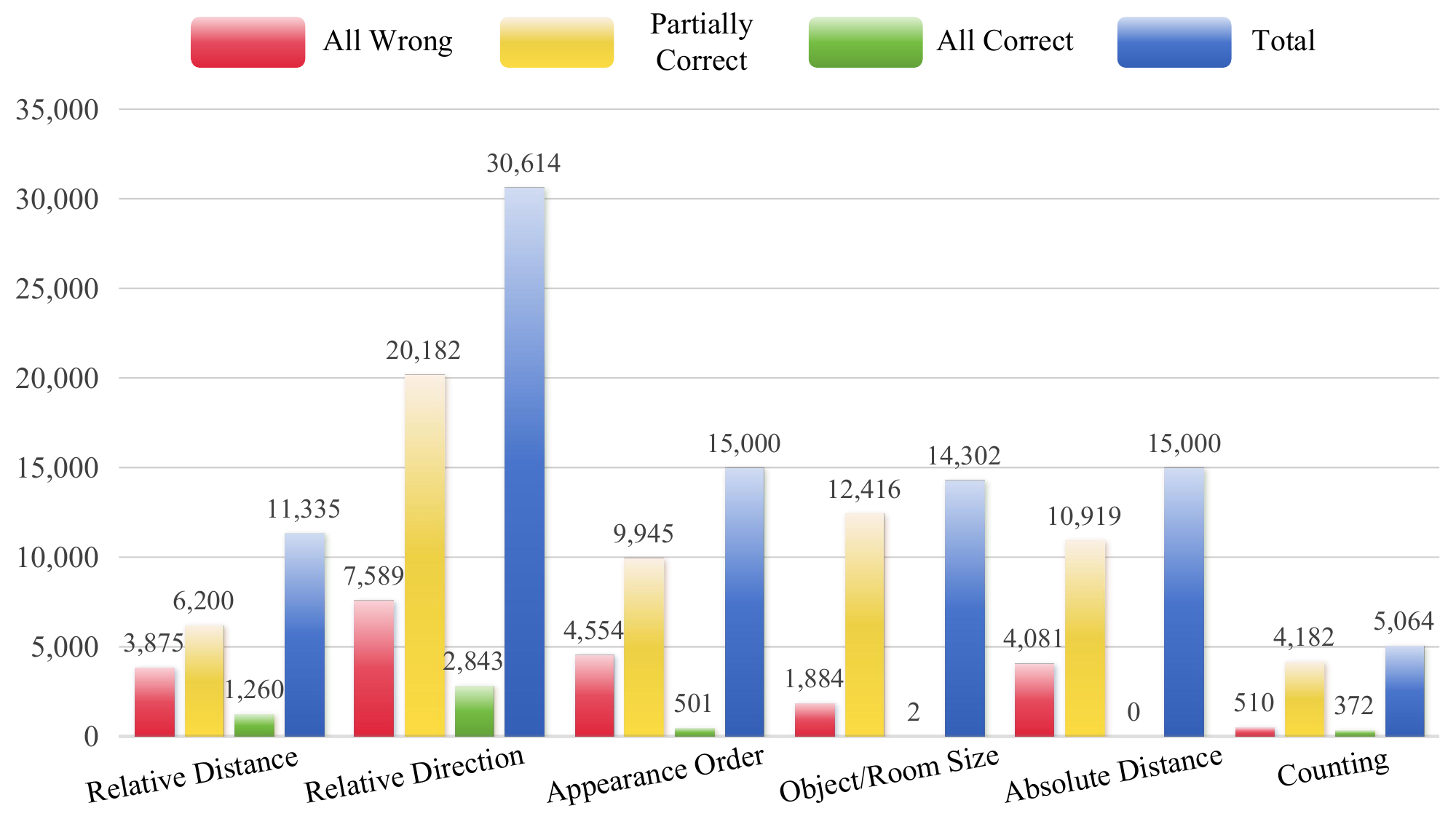}}
        \subfigure[Performance variations between pre-sampling and post-sampling.]{
            \label{Fig:data_sampling2}
            \includegraphics[width=.52\linewidth]{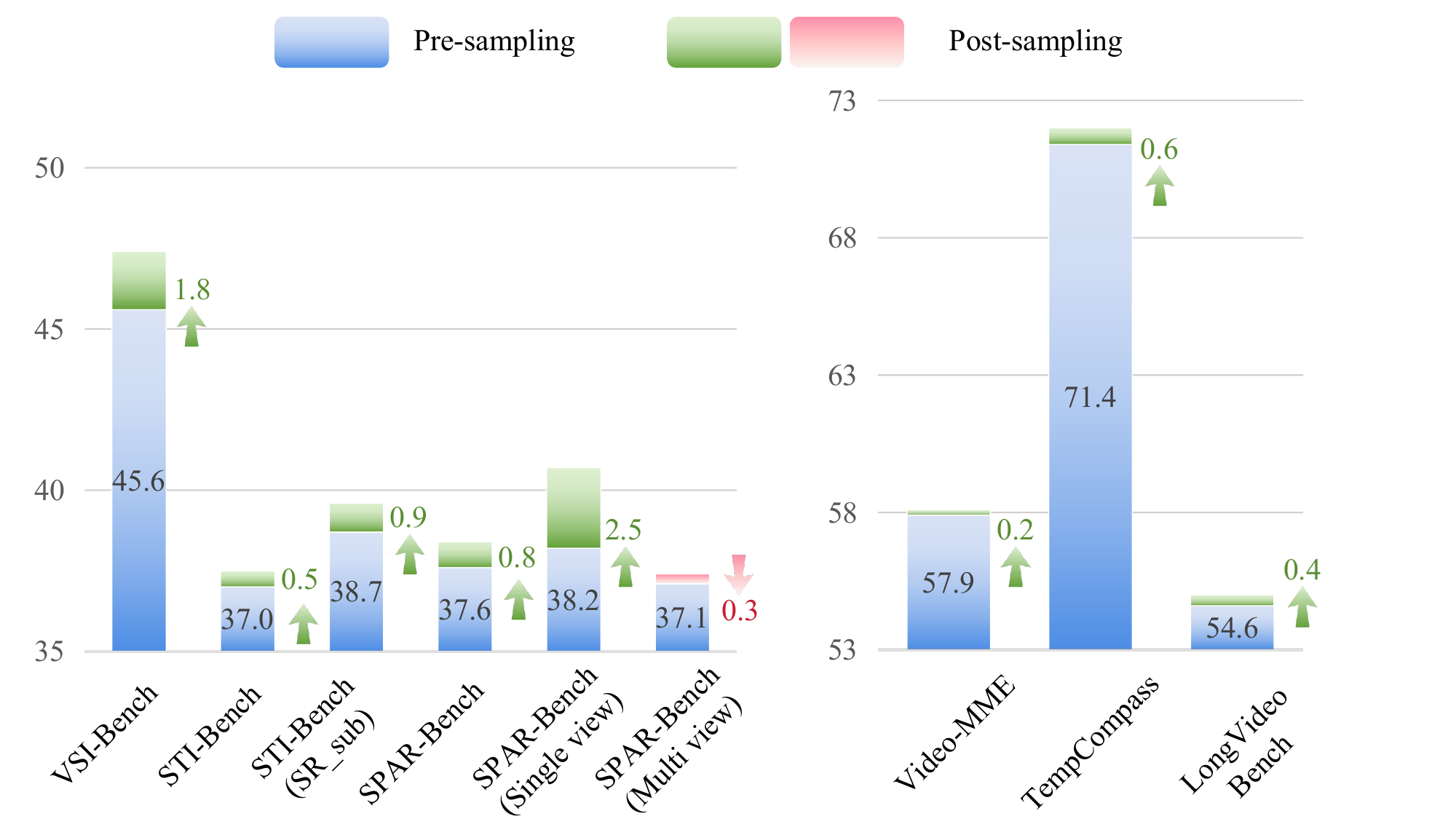}}
        \caption{Sample distribution and Performance variations between pre-sampling and post-sampling.}
      \label{Fig:data_sampling}
\end{figure*}

\noindent\textbf{Impact of Data Sampling on Model Performance.}
To improve training efficiency and model generalization, we conduct a sample selection strategy on the SR-91k dataset by filtering out samples deemed too easy or too difficult. Using Qwen2.5-VL-7B-Instruct to generate $8$ responses per sample, we categorize samples into all correct, partially correct, and all wrong, based on model response consistency, those all correct samples (low learning value) and all wrong samples (potential noise) are excluded. As illustrated in Figure~\ref{Fig:data_sampling1}, the remaining samples span a balanced range of task categories. Retraining SpaceR on this filtered dataset results in consistent performance gains across nearly all benchmarks, as shown in Figure~\ref{Fig:data_sampling2}. 
These findings suggest that targeted resampling enhances model training by focusing on samples with high learning utility, ultimately leading to improved generalization in both spatial reasoning and video understanding tasks. Additional analyses for the impacts of thinking, model size, and data scale on model performance are provided in Appendix~\ref{Appendix_analyses}.

\begin{table*}[ht]
    \centering
    \resizebox{\textwidth}{!}{
        \begin{tabular}{l|c|c|cc|ccc|c|c|c}
        \toprule
     & \multicolumn{1}{c|}{\multirow{3}{*}{Frames}}  & \multicolumn{6}{c|}{\textbf{Spatial Reasoning}} & \multicolumn{3}{c}{\textbf{Video Understanding}} \\ \cline{3-11}
             & &  \multicolumn{1}{c|}{\multirow{2}{*}{\textbf{VSI-Bench}}}& \multicolumn{2}{c|}{\textbf{STI-Bench}} & \multicolumn{3}{c|}{\textbf{SPAR-Bench}} & \multicolumn{1}{c|}{\multirow{2}{*}{\textbf{VM}}} & \multicolumn{1}{c|}{\multirow{2}{*}{\textbf{TC}}} & \multicolumn{1}{c}{\multirow{2}{*}{\textbf{LV}}} \\ \cline{4-8}
             & & & Overall & SR\_sub & Overall & Single-view & Multi-view & & & \\
        \midrule
            w/o-map imagination  & 32 &43.9  &34.0  &33.3  &34.1  &34.7  &33.8  &56.9 &\textbf{71.9} &52.8  \\
            w/o-general data  & 32 &\textbf{46.9}  &\textbf{37.1}&37.2  &37.1&35.5&\textbf{39.5}  &56.3 &71.0 &53.6  \\
            w/o-SR data  & 32 &26.2  &34.2   &33.9  &30.2 &30.7  &29.5  &57.4 &70.5 &52.9  \\
            \midrule
                SpaceR~\textcolor{gray}{SG-RLVR}  & 32& 45.6 & 37.0 & \textbf{38.7} & \textbf{37.6}& \textbf{38.2} & 37.1&\textbf{57.9}&71.4&\textbf{54.6}\\
        \bottomrule
        \end{tabular}
    }
    \caption{Ablation results of SpaceR~\textcolor{gray}{SG-RLVR}, where the best results are in boldface.}
    \label{Tab:ablation_exp}
\end{table*}

\begin{figure*}[htb]
\centering
\subfigure[Comparison on object relative distance task.]{
\centering
\includegraphics[width=0.9\textwidth]{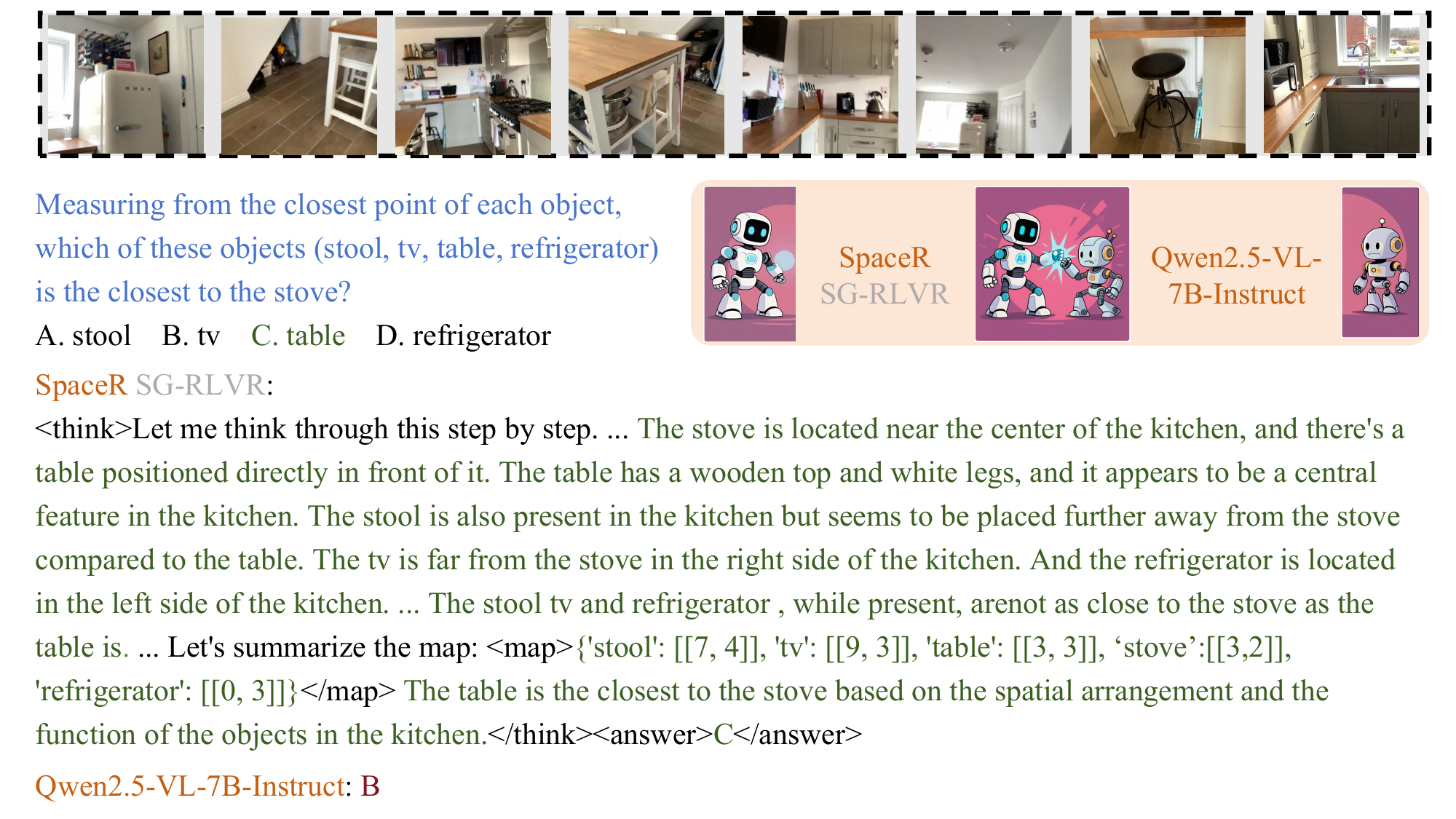}
}
\quad
\subfigure[Comparison on object counting task.]{
\centering
\includegraphics[width=0.9\textwidth]{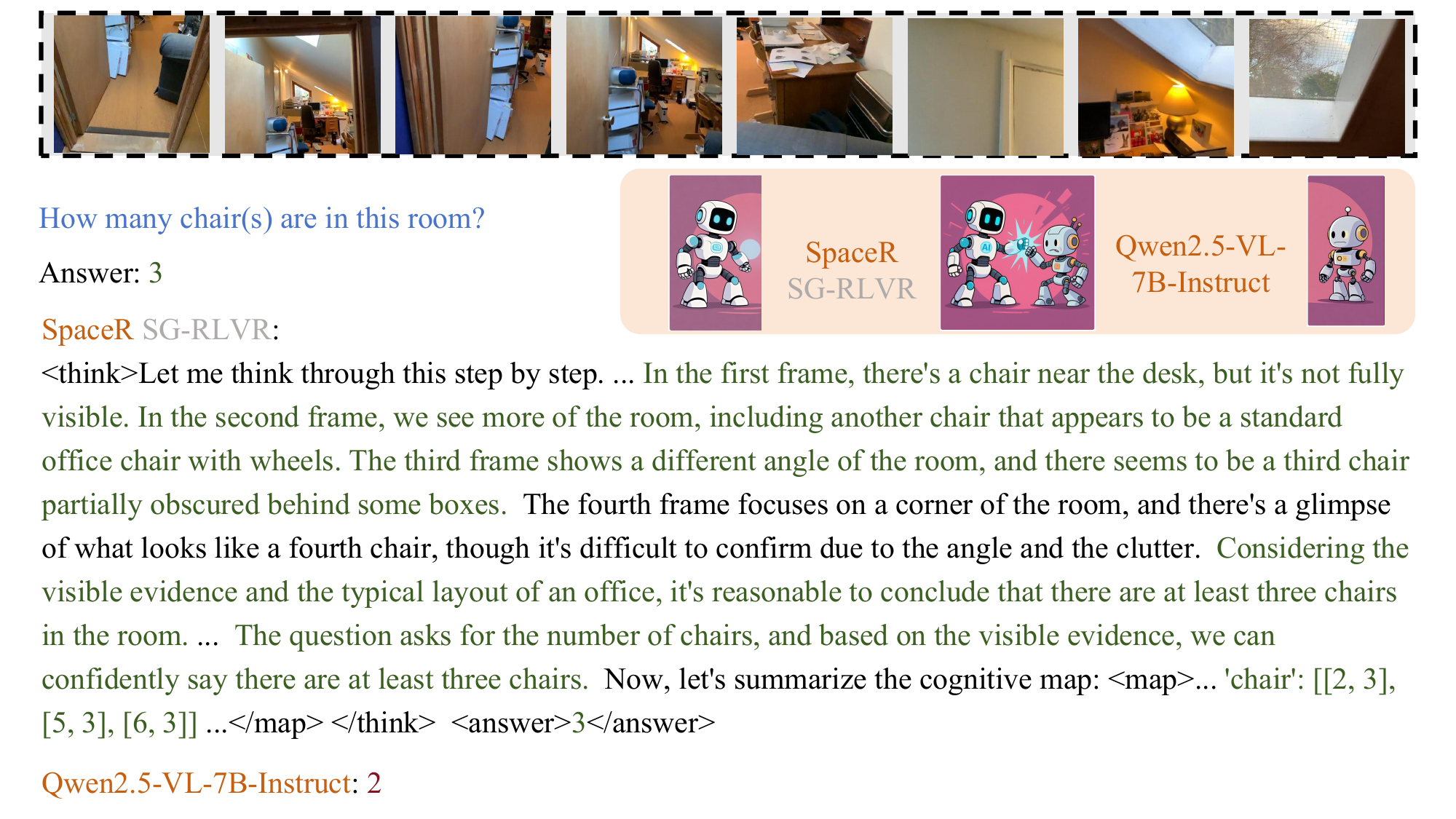}
}
\centering
\caption{Two samples from VSI-Bench~\cite{yang2024thinking}, as well as  the corresponding responses from SpaceR \textcolor{gray}{SG-RLVR} and Qwen2.5-VL-7B-Instruct~\cite{Qwen2.5-VL}.}
\label{Fig:case_study}
\end{figure*}
\subsection{Ablation Study}
To explore the contributions of individual components in our method, we introduce three variants of our SpaceR: 1) w/o-map imagination, which eliminates the map imagination mechanism in the training and inference stage. 2) w/o-general data, which excludes 60k general multimodal understanding data in the training stage. 3) w/o-SR data, which removes SR-91k data in the training process. 
The ablation results are presented in Table~\ref{Tab:ablation_exp}, based on which we have the following findings.
a) SpaceR consistently outperforms w/o-map imagination on three spatial reasoning benchmarks, which validates the advantage of map imagination mechanism to guide MLLMs in understanding spatial information. b) SpaceR exceeds w/o-general data on video understanding benchmarks, which prove the effectiveness of the $60,000$ general multimodal understanding training data. c) w/o-SR data shows significantly lower performance on spatial reasoning benchmarks compared to SpaceR, emphasizing the critical importance of SR-91k dataset in enhancing spatial reasoning capabilities.
\subsection{Qualitative Analysis}
To get an intuitive understanding on the advancement of SpaceR in video spatial reasoning, we present two cases from VSI-Bench in Figure~\ref{Fig:case_study}. In case (a), SpaceR demonstrates a clear qualitative superiority over Qwen2.5-VL-7B through structured reasoning, explicit spatial mapping, and an accurate conclusion. SpaceR correctly identifies the ``table'' as the object closest to the ``stove'', supporting its decision with a cognitive map that simulates the spatial layouts of the scene. In contrast, Qwen2.5-VL-7B relies on shallow, assumption-based heuristics and incorrectly selects the ``stool'', reflecting a lack of verifiable spatial reasoning. This emphasizes SpaceR’s enhanced spatial awareness and reasoning depth, enabled by its map imagination mechanism in the SG-RLVR framework.
Similarly, in case (b), SpaceR also beats Qwen2.5-VL-7B on a ``chair-counting'' task. By reasoning across multiple frames and accounting for partially occluded objects, SpaceR accurately concludes the presence of at least three chairs and reinforces its answer with a cognitive map. In contrast, Qwen2.5-VL-7B underestimates the count, providing a wrong answer of two chairs. 
Together, these cases prove the SpaceR’s improved ability of spatial reasoning.

\section{Conclusion} \label{main_conclusion}
In this work, we introduce SpaceR, a novel framework designed to enhance video spatial reasoning capabilities. To this end, we construct SpaceR-151k, a comprehensive dataset that includes 91k high-quality spatial reasoning QA paris (SR-91k) and 60k samples for general video understanding. Building upon this dataset, we propose Spatially-Guided Reinforcement Learning with Verifiable Reward (SG-RLVR), a novel reinforcement framework, which integrates task-specific reward functions and a map imagination mechanism to guide models in spatial layout inference and foster structured spatial reasoning.
Extensive evaluations across three spatial reasoning benchmarks and video understanding benchmarks validate the effectiveness and generalizability of SpaceR. Nevertheless, certain limitations remain. For example, the current framework lacks mechanisms for adaptively controlling the depth of reasoning, which may affect inference efficiency in practice. This is expected to be explored in our future work. We hope that SpaceR serves as a solid foundation for advancing research in video spatial reasoning and inspires further exploration into reasoning-aware training for MLLMs.

\clearpage
{
    \small
    \bibliographystyle{acl_natbib}
    \bibliography{main}
}

\medskip
\newpage
\appendix
\section{More Details for Data Construction}\label{Appendix_dataconstruction}
\begin{figure*}[h]
    \centering
    \includegraphics[width=0.95\textwidth]{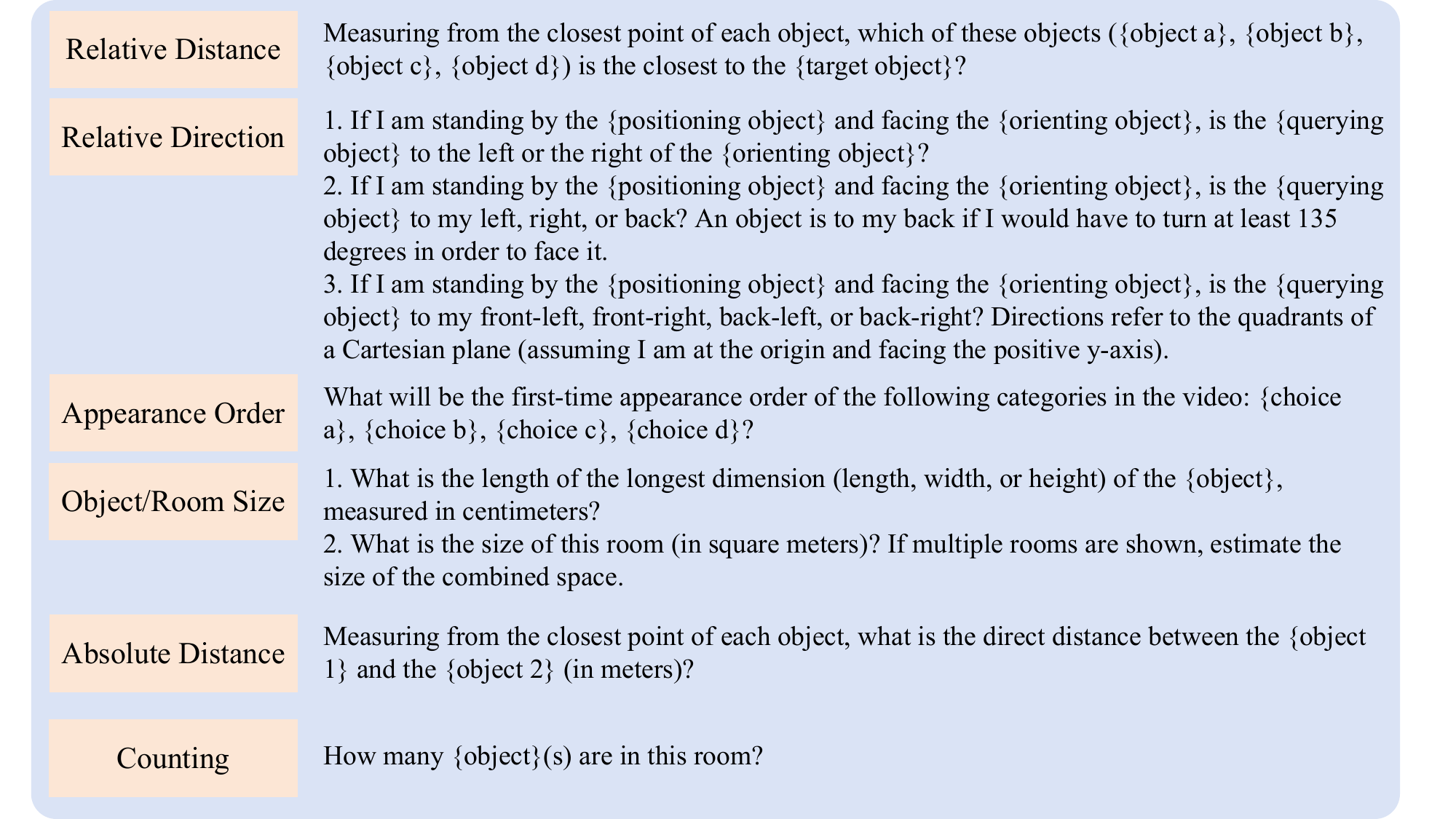}
     \caption{Question templates for QA pairs of SR-91k.}
    \label{Fig:qatemp}
\end{figure*}
\noindent\textbf{QA Generation.}
To format the generated QA pairs, we incorporate the corresponding objects into the specified question templates, which are presented in Figure~\ref{Fig:qatemp}.

\noindent\textbf{Data Filtering.}
We remove the QA pairs that involve some noisy objects (\eg ``wall'', ``floor'', and ``ceiling''). And we also drop the numerical QA pairs, where the objects are too small to identify. 
Considering VSI-Bench~\cite{yang2024thinking} is partially built on ScanNet~\cite{dai2017scannet}, we exclude the overlapped videos in our SR-91k to ensure fair evaluation.

\begin{figure*}[h]
    \centering
    \includegraphics[width=0.95\textwidth]{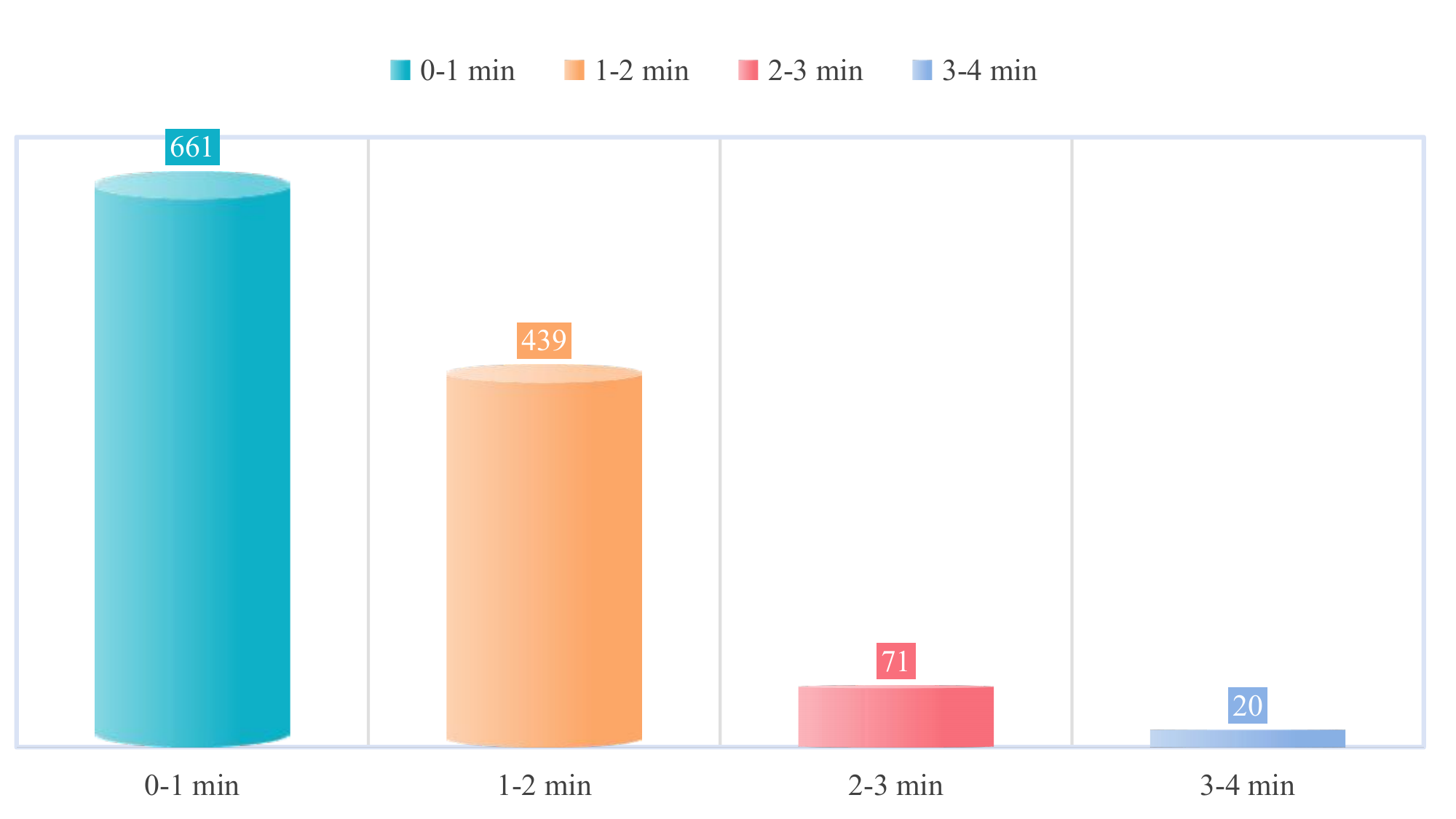}
     \caption{Video duration distribution of SR-91k.}
    \label{Fig:duration}
\end{figure*}
\noindent\textbf{Data Statistics.}
We visualize the duration distribution of videos from SR-91k in Figure~\ref{Fig:duration}.
\section{More Details for Experiment}\label{Appendix_exp}
\subsection{Benchmarks Description}\label{Appendix_bench}
\begin{itemize}
    \item \textbf{VSI-Bench}~\cite{yang2024thinking} is a comprehensive benchmark for evaluating the visual-spatial intelligence of Multimodal Large Language Models (MLLMs). It comprises over $5,000$ question-answer pairs across $288$ real-world indoor scene videos, covering diverse environments such as homes, offices, and factories, and is specifically designed to assess spatial reasoning capabilities.
    \item \textbf{STI-Bench}~\cite{li2025stibenchmllmsreadyprecise} evaluates the spatial understanding abilities of MLLMs using real-world videos spanning desktop, indoor, and outdoor scenarios. It includes eight challenging tasks, with the subset SR\_sub, which contains more than $2,000$ QA pairs across six sub-tasks (\ie Dimensional Measurement, Displacement \& Path Length,Ego-Centric Orientation, Spatial Relation, Speed \& Acceleration, Trajectory Description), being most relevant to our focus on spatial reasoning.
    \item \textbf{SPAR-Bench}~\cite{zhang2025flatland} is specifically designed to measure the spatial understanding of MLLMs. It contains over $7,000$ QA pairs covering a spectrum of tasks from basic perception to complex spatial reasoning. The benchmark is further divided into single-view and multi-view settings, allowing for comprehensive assessment across varying spatial contexts. 
    \item \textbf{Video-MME}~\cite{fu2024video} serves as a comprehensive benchmark for evaluating general video understanding in MLLMs. It includes $900$ videos and $2,700$ high-quality multi-choice questions (three per video), spanning a wide range of scenarios and tasks. We exclude the subtitles of videos in the evaluation.
    \item \textbf{TempCompass}~\cite{liu-etal-2024-tempcompass} focuses on temporal perception in MLLMs. It consists of $410$ videos and $7,540$ questions designed to evaluate models’ understanding of temporal dynamics.
    \item \textbf{LongVideoBench}~\cite{wu2024longvideobench} is a benchmark for long-context multimodal video understanding. It features $6,678$ carefully constructed multi-choice questions derived from videos of varying durations, extending up to one hour, and encompasses diverse real-world themes. We utilize the validations set of it and remove the subtitles of videos.

\end{itemize}
\subsection{Baselines Description}\label{Appendix_baseline}
\begin{itemize}
    \item \textbf{GPT-4o}~\cite{GPT-4o} is a state-of-the-art MLLM developed by OpenAI, exhibiting strong performance across a variety of vision-language tasks.
    \item \textbf{Gemini 1.5 Pro, Gemini 2.0 Flash, Gemini 2.5 Pro} are advanced MLLMs from Google's Gemini family\footnote{\url{https://aistudio.google.com}.}. These models have shown leading performance across several video understanding benchmarks (\eg Video-MME~\cite{fu2024video}, and LongVideoBench~\cite{wu2024longvideobench}). Gemini 2.0 Flash and Gemini 2.5 Pro, in particular, exhibit enhanced abilities in complex reasoning tasks.
    \item \textbf{VideoLLaMA3-7B}~\cite{damonlpsg2025videollama3} is an MLLM tailored for both image and video understanding. It adopts Qwen2.5-7B~\cite{qwen2.5} as its language backbone and integrates siglip-so400m-patch14-384~\cite{DBLP:conf/iccv/ZhaiM0B23} as the vision encoder.
    \item \textbf{LLaVA-OneVision-7B}~\cite{li2024llava} represents a strong advancement in open-source multimodal language models (LMMs), combining the Qwen2~\cite{qwen2.5} language backbone with the SigLIP~\cite{DBLP:conf/iccv/ZhaiM0B23} vision encoder. This integration pushes the performance boundaries of open LMMs, particularly in tasks requiring fine-grained visual understanding.
    \item \textbf{MiniCPM-V-2.6}~\cite{yao2024minicpm} is developed based on SigLIP-400M~\cite{DBLP:conf/iccv/ZhaiM0B23} and Qwen2-7B~\cite{qwen2.5}, and introduces enhanced capabilities for multi-image and video understanding. Its architectural improvements and task-specific design make it a competitive model for complex multimodal understanding tasks.
    \item \textbf{InternVL2.5-78B}~\cite{chen2024expanding} is a high-performing open-source MLLM that combines InternViT-6B-448px-V2\_5~\cite{chen2024internvl} as the vision encoder with Qwen2.5-72B-Instruct~\cite{qwen2.5} as the LLM backbone.
    \item \textbf{Kimi-VL-A3B-Instruct}~\cite{kimiteam2025kimivltechnicalreport} is an efficient open-source MLLM based on a Mixture-of-Experts (MoE) architecture. It incorporates the Moonlight~\cite{DBLP:journals/corr/abs-2502-16982} MoE language model and the high-resolution MoonViT~\cite{kimiteam2025kimivltechnicalreport} vision encoder.
    \item \textbf{Qwen2.5-VL-7B-Insturct, Qwen2.5-VL-72B-Insturct}~\cite{Qwen2.5-VL} are part of the Qwen2.5-VL series, which combine the Qwen2.5~\cite{qwen2.5} language model with a redesigned Vision Transformer (ViT) architecture for enhanced visual grounding and understanding.
\end{itemize}
\subsection{Hardware Usage} \label{Appendix_hardware}
Model training is conducted under $8$ L20 $80$ GiB GPUs or $4$ A800 $80$ GiB GPUs. Model is evaluated under $4$ L20 $80$ GiB GPUs.
\section{More Empirical Results and Analyses}\label{Appendix_analyses}
In this section, we supply more comparison results, and more analyses for impact of data scale on model performance.
\begin{table*}[ht]
    \centering
    \resizebox{\textwidth}{!}{
        \begin{tabular}{l|c|c|c|cc|ccc|c|c|c|c|c}
        \toprule
             & \multicolumn{1}{c|}{\multirow{2}{*}{\#Params}} & \multicolumn{1}{c|}{\multirow{2}{*}{Frames}} & \multicolumn{1}{c|}{\multirow{2}{*}{\textbf{VSI-Bench}}} & \multicolumn{2}{c|}{\textbf{STI-Bench}} & \multicolumn{3}{c|}{\textbf{SPAR-Bench}} & \multicolumn{1}{c|}{\multirow{2}{*}{\textbf{Avg. Token}}} & \multicolumn{1}{c|}{\multirow{2}{*}{\textbf{VM}}} & \multicolumn{1}{c|}{\multirow{2}{*}{\textbf{TC}}} & \multicolumn{1}{c|}{\multirow{2}{*}{\textbf{LV}}}& \multicolumn{1}{|c}{\multirow{2}{*}{\textbf{Avg. Tokens}}} \\ \cline{5-9}
             & & & & Overall & SR\_sub & Overall & Single-view & Multi-view & & & \\
        \midrule
          \rowcolor{lightgray}Qwen2.5-VL-3B-Instruct~\cite{Qwen2.5-VL}&3B&32&-&-&-&-&-&-&-&-&-&-&-\\
         \quad + \textcolor{gray}{non-think}&-&-&26.7&36.7&37.5&25.4&25.3&25.5&-&52.4&65.7&49.7&-\\
           \quad + \textcolor{gray}{think}
               &-&-&25.9~\textcolor{jweired}{ \small($\downarrow$ 0.8)} &34.3~\textcolor{jweired}{ \small($\downarrow$ 2.4)} &37.2~\textcolor{jweired}{ \small($\downarrow$ 0.3)} &26.8~\textcolor{jweigreen}{ \small($\uparrow$ 1.4)}&26.6~\textcolor{jweigreen}{ \small($\uparrow$ 1.3)}&26.9~\textcolor{jweigreen}{ \small($\uparrow$ 1.4)}&66.9&52.0~\textcolor{jweired}{ \small($\downarrow$ 0.4)}&63.2~\textcolor{jweired}{ \small($\downarrow$ 2.4)}&49.0~\textcolor{jweired}{ \small($\downarrow$ 0.7)}&0.3
               \\
               \rowcolor{lightgray}SpaceR-Tiny \textcolor{gray}{SFT}& 3B & 32 &-&-&-&-&-&-&-&-&-&-&-\\
              \quad + \textcolor{gray}{non-think}& - & -&34.8&33.0&36.5&24.8&24.5&24.9&53.4&63.8&50.7&-&-\\
               \rowcolor{lightgray}SpaceR-Tiny \textcolor{gray}{SG-RLVR}& 3B & 32 &-&-&-&-&-&-&-&-&-&-&-\\
             \quad + \textcolor{gray}{non-think}& - & -&40.5&36.6&38.7&30.1&30.7&29.6&-&52.9&66.4&50.1&-\\
            \quad + \textcolor{gray}{think}& - & -&41.2~\textcolor{jweigreen}{ \small($\uparrow$ 0.7)}&37.8~\textcolor{jweigreen}{ \small($\uparrow$ 1.2)}&40.1~\textcolor{jweigreen}{ \small($\uparrow$ 1.4)}&30.9~\textcolor{jweigreen}{ \small($\uparrow$ 0.8)}&31.4~\textcolor{jweigreen}{ \small($\uparrow$ 0.7)}&30.6~\textcolor{jweigreen}{ \small($\uparrow$ 1.0)}&274.2&51.6~\textcolor{jweired}{ \small($\downarrow$ 1.3)}&65.4~\textcolor{jweired}{ \small($\downarrow$ 1.0)}&49.4~\textcolor{jweired}{ \small($\downarrow$ 0.7)}&237.1\\
              \midrule
          \rowcolor{aliceblue}Qwen2.5-VL-7B-Instruct~\cite{Qwen2.5-VL} & 7B & 32 &-&-&-&-&-&-&-&-&-&-&-
              \\
              \quad + \textcolor{gray}{non-think}
              &-&-& 34.4 & 34.5  & 32.3 & 33.8 & 36.9 & 31.8 &-& 56.3 & 71.1& 53.5&- \\
            \quad + \textcolor{gray}{think}& - & - &30.2~\textcolor{jweired}{ \small($\downarrow$ 4.2)}&33.2~\textcolor{jweired}{ \small($\downarrow$ 1.3)}&34.4~\textcolor{jweigreen}{ \small($\uparrow$ 2.1)}&31.6~\textcolor{jweired}{ \small($\downarrow$ 2.2)}&31.2~\textcolor{jweired}{ \small($\downarrow$ 5.7)}& 31.8~\textcolor{yellowish}{ \small(- 0.0)}&104.0 &54.0~\textcolor{jweired}{ \small($\downarrow$ 2.3)} &68.1~\textcolor{jweired}{ \small($\downarrow$ 3.0)} &46.6~\textcolor{jweired}{ \small($\downarrow$ 6.9)}&68.6\\

            \rowcolor{lightblue}SpaceR \textcolor{gray}{SG-RLVR}& 7B & 32 &-&-&-&-&-&-&-&-&-&-&-\\
            \quad + \textcolor{gray}{non-think}& - & -&45.0&36.7&34.8&36.1&36.2&36.0&-&57.9&71.4&54.6&-\\
            \quad + \textcolor{gray}{think}& - & - & {45.6}~\textcolor{jweigreen}{ \small($\uparrow$ 0.6)}  & {37.0}~\textcolor{jweigreen}{ \small($\uparrow$ 0.3)}  & {38.7}~\textcolor{jweigreen}{ \small($\uparrow$ 3.9)} & {37.6}~\textcolor{jweigreen}{ \small($\uparrow$ 1.5)} & {38.2}~\textcolor{jweigreen}{ \small($\uparrow$ 2.0)} & {37.1}~\textcolor{jweigreen}{ \small($\uparrow$ 1.1)}&345.6 &56.4~\textcolor{jweired}{ \small($\downarrow$ 1.5)}  &70.0~\textcolor{jweired}{ \small($\downarrow$ 1.4)}   &51.7~\textcolor{jweired}{ \small($\downarrow$ 2.9)} &265.7  \\
        \bottomrule
        \end{tabular}
    }
    \caption{We compare non-think and think modes of SpaceR-Tiny, Qwen2.5-VL-3B-Instruct, \colorbox{lightblue}{SpaceR}, \colorbox{aliceblue}{Qwen2.5-VL-7B-Instruct} on spatial reasoning benchmarks (VSI-Bench, STI-Bench, and SPAR-Bench), and video understanding benchmarks: \textbf{VM} (Video-MME), \textbf{TC} (TempCompass), and \textbf{LV} (LongVideoBench). \textcolor{gray}{non-think} means directly outputting answers, while \textcolor{gray}{think} refers to explicitly engaging thinking process during inference. \textbf{Avg. Tokens} denotes to the tokens number of thinking process in the generated responses. 
    } 
    \label{Tab:app_exp}
\end{table*}

\subsection{Impact of thinking on Model Performance}\label{Appendix_com_res}
To assess the effect of explicitly engaging the thinking process during inference, we compare model performance under two modes: \textit{non-think}, which outputs answers directly, and \textit{think}, which includes a structured reasoning process. As shown in Table~\ref{Tab:app_exp}, models not explicitly trained to reason, such as Qwen2.5-VL-3B-Instruct, Qwen2.5-VL-7B-Instruct, and SpaceR-Tiny \textcolor{gray}{SFT}, exhibit a significant performance drop across most benchmarks in \textit{think} mode. This degradation is expected, since these models lack sufficient reasoning capability and often generate shorter, less informative reasoning traces, which could be reflected by their lower average token counts. In contrast, our SpaceR \textcolor{gray}{SG-RLVR} demonstrates consistent gains on spatial reasoning benchmarks when utilizing the \textit{think} mode. This indicates that our SG-RLVR method helps the model develop useful reasoning strategies. However, the same trend does not extend to video understanding tasks, where SpaceR shows a slight performance drop in \textit{think} mode. This suggests that for such tasks, unnecessary or inaccurate reasoning may introduce noise, hindering model predictions.
This raises a critical question: \textit{Does Thinking Really Helps?} We hypothesize two contributing factors: 1) While our SG-RLVR framework effectively optimizes for accurate answers, it does not provide direct supervision for the reasoning process itself, leading to suboptimal or inconsistent reasoning traces; 2) For some tasks like video understanding, reasoning may be redundant or even misleading, especially when the question can be answered directly and incorrect reasoning introduces spurious information. These observations highlight a valuable direction for future work: enabling models to decide \textit{when to think} and \textit{how to think accurately}. Developing mechanisms to supervise or adaptively trigger the reasoning process could further enhance performance and interpretability in MLLMs.
\subsection{Impact of Model Size on Performance.} \label{Appendix_modelsize}
To examine the scalability of our SG-RLVR framework across different model sizes, we fine-tune both Qwen2.5-VL-3B-Instruct and Qwen2.5-VL-7B-Instruct~\cite{Qwen2.5-VL} on the SpaceR-151k dataset using both supervised fine-tuning (SFT) and our proposed SG-RLVR method. As shown in Table~\ref{Tab:app_exp}, the base models exhibit limited spatial reasoning capabilities, with noticeable performance drops in the \textit{think} mode, where they, especially Qwen2.5-VL-3B-Instruct, struggle to produce coherent reasoning and often revert to direct answer outputs. Although SFT leads to modest gains on benchmarks such as VSI-Bench, it fails to enhance generalization and does not substantially improve reasoning ability.
In contrast, our SpaceR-Tiny \textcolor{gray}{SG-RLVR} and SpaceR \textcolor{gray}{SG-RLVR} consistently outperform the base models and their SFT counterparts across multiple spatial reasoning benchmarks, particularly in the \textit{think} mode. Meanwhile, they show promising generalizability on video understanding benchmarks. These results confirm both the effectiveness and scalability of the SG-RLVR framework in enhancing spatial reasoning across model sizes.

\begin{figure*}[htb]
       \centering
        \subfigure[Spatial Reasoning Benchmarks.]{
            \label{Fig:data_scale1}
            \includegraphics[width=.49\linewidth]{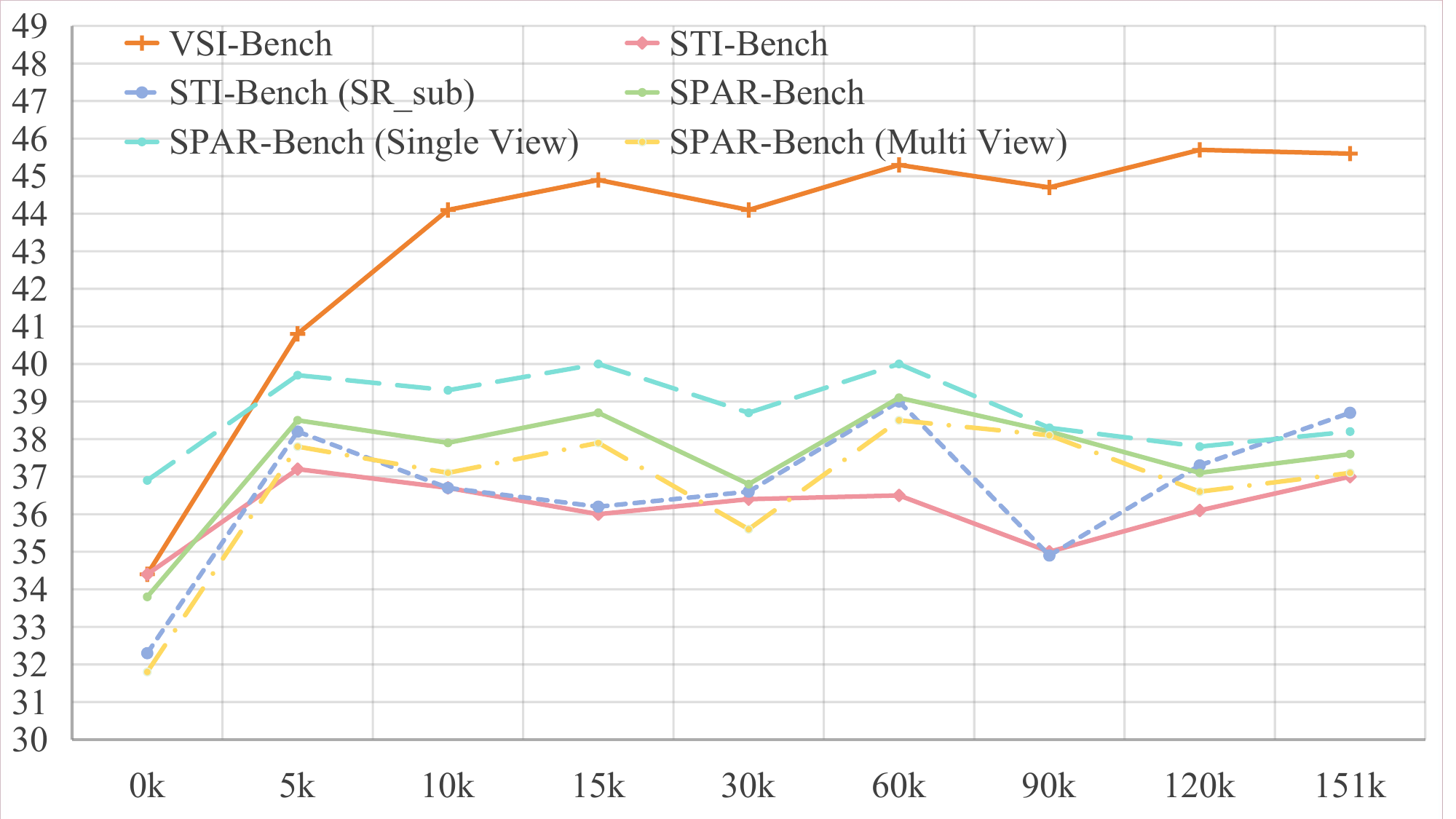}}
        \subfigure[Video Understanding Benchmarks.]{
            \label{Fig:data_scale2}
            \includegraphics[width=.49\linewidth]{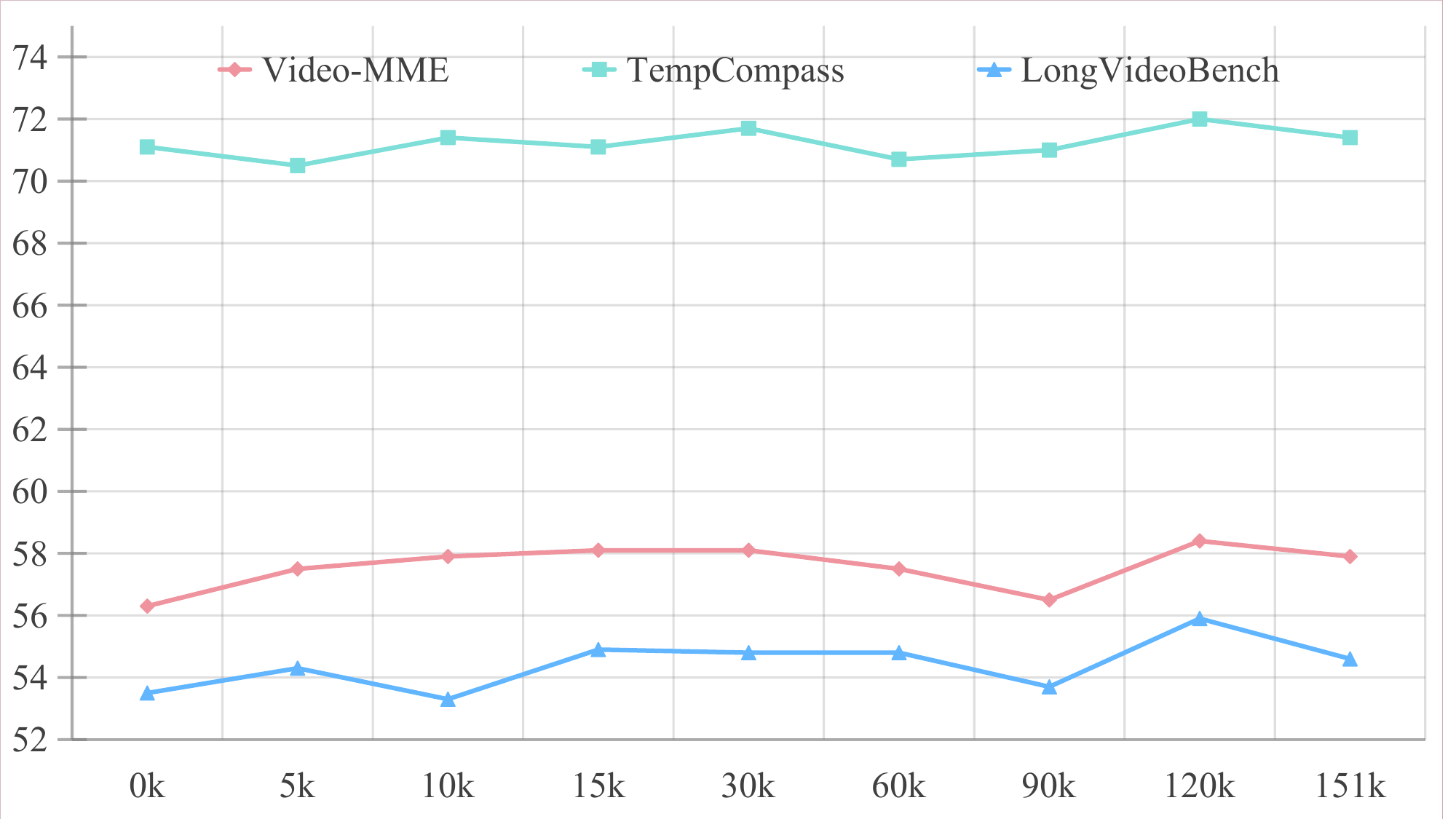}}
        \caption{Performance variations with progressively increasing data scale on spatial reasoning and video understanding benchmarks.}
      \label{Fig:data_scale}
\end{figure*}

\subsection{Impact of Data Scale on Model Performance.} \label{Appendix_datascale}
To examine the relationship between data scale and model performance, we train SpaceR on six progressively larger subsets of the SpaceR-151k dataset. The results, presented in Figure~\ref{Fig:data_scale}, reveal two key observations. First, SpaceR demonstrates notable performance improvements on spatial reasoning benchmarks such as VSI-Bench and SPAR-Bench (Single View) even with small-scale subsets (\eg 5k–15k), highlighting the strong data efficiency of our SG-RLVR training paradigm. Second, while performance generally plateaus or slightly fluctuates as the data scale increases to 30k, a substantial jump is observed when training on the full dataset (151k), especially for VSI-Bench, where accuracy surpasses $45\%$. These findings suggest that our method not only benefits from larger training sets but also exhibits strong generalization from limited supervision, underscoring its effectiveness in both low-resource and full-scale settings.

\section{Documentation and Licensing}\label{Appendix_license}
The SpaceR-151k dataset includes annotations in JSONL format, along with associated images and videos. The SpaceR model adopts the same architecture as Qwen2.5-VL-7B-Instruct. Both the dataset and the model are released under the CC BY-NC 4.0 license\footnote{\url{https://creativecommons.org/licenses/by-nc/4.0/}} and are intended for academic research purposes only.
Additionally, the ScanNet~\cite{dai2017scannet} dataset has been released under MIT license.

\end{document}